# Particle Swarm Optimization for Quantum Circuit Synthesis: Performance Analysis and Insights

*Mirza Hizriyan Nubli Hidayat[1], Tan Chye Cheah[1]

[1]School of Computer Science, University of Nottingham Malaysia

**Abstract:**

This paper discusses how particle swarm optimization (PSO) can be used to generate quantum circuits to solve an instance of the MaxOne problem. It then analyzes previous studies on evolutionary algorithms for circuit synthesis. With a brief introduction to PSO, including its parameters and algorithm flow, the paper focuses on a method of quantum circuit encoding and representation as PSO parameters. The fitness evaluation used in this paper is the MaxOne problem. The paper presents experimental results that compare different learning abilities and inertia weight variations in the PSO algorithm. A comparison is further made between the PSO algorithm and a genetic algorithm for quantum circuit synthesis. The results suggest PSO converges more quickly to the optimal solution.

## 1. Introduction

The advent of quantum computing has ushered in a new era of computational potential, promising solutions to classically intricate problems. Many predict the importance of quantum computing to increase in the coming years. This has sprung many investments into the world of quantum computers, creating a larger availability of quantum computers to the general public [1]. An increase of this level requires further development in quantum algorithms. The superiority of quantum computing over classical computing has been proven since its earlier stages through algorithms such as Shor's factoring algorithm [2] and Grover's search algorithm [3]. Shor's algorithm showcases the prime factors of an $n$-digit number in polynomial time, while the best-known classical factoring runs at $O(2^{n^{1/3} log(n)^{2/3}})$. Grover's database search finds within an unsorted list of $n$ items in just $O(\sqrt{n})$, whilst most classical search runs at $O(n)$.

Recent research on quantum algorithms has focused more towards implementing a quantum version of a classical algorithm. One of which results in many quantum counterparts of evolutionary algorithms being proposed such as the Genetic Quantum Algorithm (GQA) [4]. Some research on quantum-inspired evolutionary algorithms for classical computing has also been garnering renewed attention [5]. These developments enable the design and implementation of more efficient and powerful computing with an overarching goal to solve more complex problems. Algorithms that leverage the unique properties of quantum systems are just a step forward towards that goal. In pursuit of this goal, this work intends to experiment with the capabilities of a classical evolutionary algorithm from a quantum-inspired perspective.

This paper offers and studies a particle swarm optimization (PSO) application to synthesise quantum circuits, that when evaluated, solves the MaxOne problem. Whilst the end result is a classical algorithm, its development was focused on a quantum representation in PSO to which a desirable and optimised quantum circuit could be developed and evaluated. The exact classification for the MaxOne problem is unknown, but it has been used as a trivial problem for evolutionary frameworks. The organisation of this paper is as follows; Section 2 previews a brief concept of PSO. Section 3 focuses on the PSO implementation for Quantum Circuit generation, and the remaining sections further analyse this application with comparisons and experimental results. This work utilises OpenQASM and Qiskit Aer to experiment and represent the synthesised quantum circuit.

## 2. Related works and Basis of Research

The use of evolutionary algorithms for circuit synthesis, or any sort of electronic or logical generation, is not uncommon and with works done throughout the years. This section discusses a few of those works, leading on towards an overview that becomes the basis of research within this work and its experimentation. Related works such as [6] and [7] become the main basis and inspiration for the work done in this paper, yet other related works in areas such as evolutionary electronics [8, 9] and even designing combinational logic circuits using genetic algorithms [10, 11]. Most of the works discussed here focus on utilising evolutionary algorithms to generate complex systems, in the form of a circuit, combinational or electronics.

Stoica et al. [8] discuss the evolutionary synthesis of analogue and digital circuits designed at the transistor level, focusing on achieving the desired functional response. The work explains a process from experimentation to implementation that requires the inclusion of evaluation techniques not commonly used in conventional design. Two of many techniques demonstrated within the paper were Cartesian Genetic programming (CGP), which was shown to outperform conventional synthesis tools in terms of circuit size reduction, and EAs. The EAs implemented in the work, such as Genetic Algorithm (GA) and PSO, were successful in designing and optimising digital circuits. These techniques were demonstrated through the fabrication of circuits in a 0.5 $\mu$m CMOS technology. Phelps et al. [9] presented Anaconda, a simulation-based synthesis tool for analogue circuits that utilises a stochastic search pattern. Essentially, the work developed a numerical search algorithm that combines evolutionary algorithms in its process, with pattern searching for synthesising the analog circuits. The work aims to find optimised circuit topologies and component values by exhaustively enumerating possible solutions, using evolutionary search methods. It produces novel results that are more effective than prior works and do not restrict synthesised topologies to known structures.

Coello et al. [10], one of the earlier works on evolutionary circuit synthesis, introduces a method based on a Genetic Algorithm (GA) approach to design combinational logic circuits. The work describes important issues to consider when solving the circuit design problem, such as the representation scheme, encoding function, and definition of the fitness function. They presented several circuits derived from their system under various assumed constraints, such as a maximum number of allowable gates and the types of available gates. They showed that their representation approach, compared to standard binary encoding, produces better performance and optimised circuits in terms of the quality of solution and the speed of convergence.

Reis et al. [6] presented a PSO-based method of logic circuit synthesis. It explores the usage of swarms and collective intelligence for solving function optimization problems in circuit synthesis. It mentions a comparison of the method with two other EAs, namely the Genetic Algorithm and the Memetic Algorithm. The results show the statistical characteristics of the PSO algorithm in comparison to GA and MA. Ong et al. [7] showed one of the first Genetic Algorithm implementations for a quantum circuit synthesis, rather than a classical circuit synthesis. The genetic algorithm interacts directly with Qiskit Aer for evaluation and presents the results of the synthesis based on the MaxOne problem. [7] presented to be one of the first to utilise an evolutionary algorithm on quantum circuit synthesis, and this paper extends upon their work by utilising a different EA, specifically PSO, for quantum circuit synthesis. The inspiration for utilising PSO for circuit generation comes from [6].

## 3. Particle Swarm Optimisation

a. **An Introduction**

Particle Swarm Optimisation emerged as an evolutionary algorithm inspired once more by nature, more specifically in the nature of swarm intelligence. A concept whereby the collective behaviour of a population that is interacting locally creates a globally coherent pattern. This inspiration resulted in an abstraction into the mathematical field. The exact cornerstones of PSO encapsulate swarm intelligence principles, such as the proximity and quality principle, and social concepts [12]. Eberhart and Kennedy [13] were the first to introduce an algorithm replicating the concept of swarm intelligence, dubbed PSO, an optimisation method discovered through simplified social model simulations of bird flocking and fish schooling. Before we get into the quantum circuit representation of PSO parameters, a baseline understanding of the classical PSO parameters may be necessary.

b. **The Parameters**

A "particle" in PSO refers to a member in the population and is subject to a velocity and position, which moves it towards a better mode of (or optimal) behaviour. In comparison towards other evolutionary algorithms, particles are used instead of genetic operators. Each particle will adjust its velocity and position (flying) according to its own and other members' experiences. A particle's position in the problem hyperspace becomes a model of a possible solution, and a particle moves throughout this hyperspace. The position of each particle is determined (classically) by the vector $x_i \in R^n$ and its movement by the velocity of the particle $v_i \in R^n$ [14], as shown in (1). The manipulation of this can also be explained through (2).

$$x_i(t) = x_i(t-1) + v_i(t) \ldots\ldots (1)$$
$$v_i(t) = v_i(t-1) + c_1 rand()\,(p_i - x_i(t-1)) + c_2 rand()(g_i - x_i(t-1)) \ldots\ldots (2)$$

The method of learning in PSO is based on a population's general experience, hence the information available to them are each particle's own experience and the knowledge of performance of other individuals in the population. These two factors can vary and provide differences in aspects of learning. This variance is defined by the constants $c_1$ and $c_2$, along with the $rand()$ demonstrating a random number of uniform distribution between [0.0, 1.0]. The set of $X_i = \{x_0, x_1, \cdots, x_i\}$ represents a position for the the particle $i$. The $p_i$ denotes the best previous position (the position of the particle which gives the best fitness value) and $g_i$ represents the global best previous position (the global best position of the population. These two sections of the velocity change allow for a linear attraction of a particle's velocity to the best position found by the given particle and to the best position found by any particle.

Additionally, a new parameter could be introduced into the equation for PSO. This new parameter would allow a balance between the global and local search abilities of a particle within the algorithm. It is known as the inertia weight and is often represented with a $w$. The addition would result in equation (3)

$$v_i(t) = w \cdot v_i(t-1) + c_1 rand()\, (p_i - x_i(t-1)) + c_2 rand()(g_i - x_i(t-1)) \ldots (3)$$

This inertia weight influences how much of the previous velocity is taken in as a factor towards the new calculated velocity. Hence, this would affect the searching ability of a particle. A larger weight would result in more global searching, whilst a smaller weight would result in more local search. This is because a larger weight would result in greater distances of travel (due to a higher constant) for a given particle, and vice versa.

There are various other parameters such as the constriction coefficient $k$, but for the sake of simplicity, this implementation only utilises the above-defined PSO parameters. It should be noted that the quantum implementation of the PSO algorithm may look different and vary, but the purpose of this work is to ensure the same essence of PSO learning and algorithm, whilst utilising quantum circuit representation.

c. **The Algorithm**

Like various evolutionary algorithms, PSO is simple in concept and in implementation. As noted to be inspired by nature's swarming characteristics, the PSO algorithm follows the same concept. A swarm of particles moves and updates their relatives position at each iteration, and updating the PSO algorithm to perform the search process to the optima. PSO also holds a population of candidate solutions (particles) that is then evaluated and higher fitness-evaluated particles' position values are recorded. The candidate solution evolves and changes due to the velocity changes based on the local and global best values of the population. This ensures that the resultant (or proceeding iteration) moves towards a global optima. Figure 1 shows a simple flowchart noting the basic PSO algorithm.

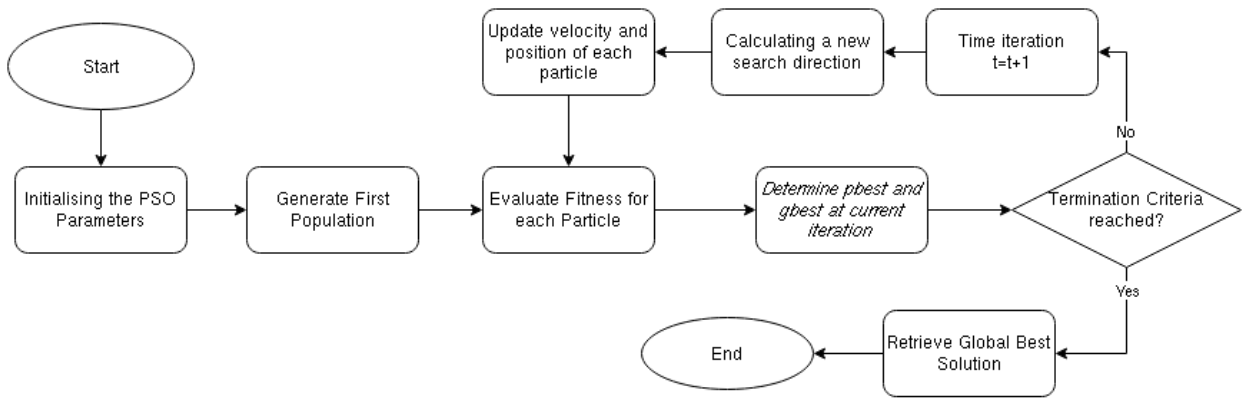

Figure 1

### d.  Variants and Experimentation

There are variants of a PSO algorithm, originally, PSO is designed to solve binary problems. Hence, for solving more real-world problems, a discrete verison was designed to solve event-based problems. Some variants of PSO simply differ in implementation and experimentation through variations in parameters used. A more prominent example of this is through different Inertia Weight considerations or the termination criteria for the algorithm. There are three inertia weight considerations that can be done for a PSO implementation, the conventional Constant Inertia Weight (CIW), a Time-Varying Inertia Weight (TVIW) and inertia weight defined through a function of the local and global best (GLbestIW). Time-Varying Inertia Weight initially proposed in [15] to improve the general performance of PSO, and GLbestIW method was first introduced in [16].

## 4.  PSO-based Quantum Circuit Synthesis

The concept behind utilising evolutionary algorithms for circuit synthesis has long been backed by works such as [17]. This paper is partly inspired by an area of research where evolutionary computing is applied to the design of electronic circuits and systems, known as Evolutionary Electronics (EE) [18]. In this area, instead of using human-designed systems and circuits, it employs search algorithms to develop implementations of systems and circuits. Several papers have offered evolutionary algorithms for combinatorial logic circuits, specifically a genetic algorithm [19]. This section dives into the representations used to complete a PSO framework for quantum circuit synthesis, from the encoding to its parameters.

### a.  Problem Statement

Before diving into the direct representations, we will briefly describe the problem description. We intend to develop a PSO algorithm capable of generating (evolving) a quantum circuit, with the final goal of solving the trivial MaxOne problem through an evaluation of the circuit. The classical MaxOne binary string entails a goal to maximise the number of 1s in a collection of binary digits. In this case, however, we implement a quantum equivalence of the MaxOne binary where instead of a predefined '0' or '1', the problem would be set to maximise the probability in the state vector "11111" for a 5 quantum bit (qubit) circuit. In this work, we utilise Qiskit Aer [20] to evaluate the synthesised circuit and retrieve its state vectors of all bit string probabilities.

### b.  Encoding

As qubit states in a circuit can vary by a huge margin, we set the Hadamard gate to be the first gate in a particle's circuit and the PSO would evolve the circuit from this Hadamard gate. The Hadamard gate is applied to all the qubits to set them in a superposition state, which ensures an equal initial probability (of '0' or '1') in each qubit [21,22].  The PSO implementation described in this work follows the generalised model of the PSO algorithm (Section 3. c) [18] with modifications in the method of representing the particles, their properties, and parameters.

In this PSO scheme, the encoding of a quantum circuit is in the form of a list of instructions within the OpenQASM program (to simulate and evaluate the circuit). A list is a simple data structure, which would help in the manipulation of the particle's velocity and position, and more so managing the size of the quantum program to prevent the algorithm from generating an infinitely large program. A particle's encoding could be further broken down through its gene representation; a gene in this representation is each instruction listed within the list. In OpenQASM [23], a quantum instruction is divided into the quantum gate to be applied (and its needed parameters) and the target qubit. This structure is further described in Figure 2. This encoding helps in the initial randomisation operation within PSO as the randomisation will have a direct effect on the gates and the parameters within the program.

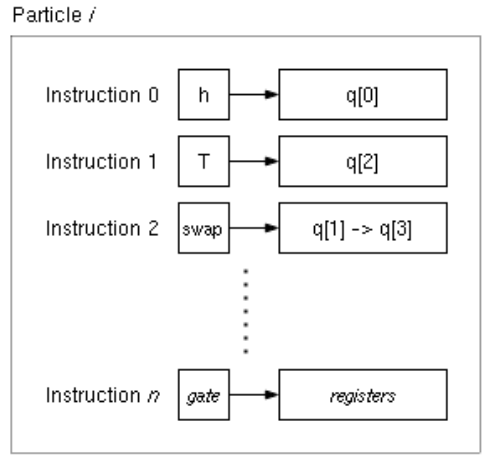

Figure 2

c. **Parameters and their Representations**

As initially stated in Section 3.b, the initial population of particles (circuits) holds a Hadamard gate instruction applied to all qubits within the circuit. The particles then go through a randomisation operation, an essence of stochasticity, within the first iteration. This enables the particles to have a velocity and position. When referring back to the classical PSO, the position of a particle can be easily represented using a vector denoting the position in a $d$-dimensional space. In the context of this quantum circuit synthesis, the position of a particle represents the configuration of gates and its parameters, more specifically the current list of instructions encoded within a particle. With this representation of a particle's position as context, we could assume that the velocity of a particle represents the changes that can be or is made to the circuit during each iteration, shown in equation (4). In this way, each candidate solution (circuit) could fly through the problem space. With this, a slight modification is required in interpreting the equations (1) and (2) (noted in Section 3. b).

The main aspect of change that the PSO is targeting, is the list of instructions or gates within the circuit; a finite list. Hence, some factors within equations (1) and (2), such as the constants $c_1$ and $c_2$ is not able to exceed the length of the list of gates. It should also be noted that the resultant representations of both the velocity and position of a particle will also be a list. With this context, the $rand()$ function would be obsolete as it returns a random constant between 0 and 1. It is not possible to multiply a constant with a list of instructions, hence the $rand()$ function would be represented by a $random.sample()$ function (built into Python [24]). This function takes in a constant and a list; returning a list of the constant's size populated randomly from the original list. This function would ensure the essence of randomness as a factor in the algorithm's learning. We can then simplify equation (2) by substituting various variables into a more simplistic form, in equation (5).

$$V_i(t-1) = x_i(t-1) - x_i(t-2) \quad \ldots\ldots (4)$$
$$V_i(t) = V_i(t-1) + rand.sample(pbest_i, c_1) + rand.sample(gbest_i, c_2) \ldots (5)$$
$$x_i(t) = x_i(t-1) + V_i(t) \quad \ldots\ldots (6)$$

- $c_1 < len(pbest_i)$
- $c_2 < len(gbest_i)$

The above is a simplified equation that produces the new velocity and the new position for a particle $i$ at an iteration. $pbest_i$ represents the list of gates that produced a particle's best fitness value at a given iteration (local best), and $gbest_i$ represents the list of gates that produced a population's best fitness value (global best). These two representations are a list of instructions or gates as well, hence the resulting new velocity for a particle $i$ is a list of instructions.

Additionally, and to further explore and analyse this PSO implementation, we have added an OpenQASM quantum circuit representation for the inertia weight. As previously explored in Section 3. b, classical PSO uses the inertia weight to control the local and global search capabilities of a particle at a given iteration. In this work's context, we focus on a representation that achieves a similar control. We utilise a given constant $w_t$ for a given iteration $t$ which is used as the size of a random sampling taken from the previous velocity of particle $i$. This utilisation of a constant in this implementation is similar to that of the constants $c_1$ and $c_2$. We can add this parameter into equation (6) which results in the following:

$$V_i(t) = rand.sample(V_i(t-1), w_t) + rand.sample(pbest_i, c_1) + rand.sample(gbest_i, c_2)$$

### d. Fitness Evaluation

There are various ways of fitness evaluation for a PSO implementation, with [6] presenting a fitness evaluation for circuit synthesis as well. The fitness evaluation allows the algorithm to assess and guide the search towards a global minimum of the objective [25]. This evaluation is used to assess the best particles within the population, thus resulting in a global best particle (circuit) at a given iteration. The evaluation also enables the algorithm to update a particle's position and velocity according to what's best and move towards the objective. Within the context of quantum circuit synthesis, [7] provides a basis for a fitness evaluation of a quantum circuit for the MaxOne problem. Hence, this PSO implementation will follow the same fitness evaluation. Further reasoning can be found in Section 2. The evaluation in [7] provides two approaches that use the probabilities computed by Qiskit Aer when a circuit (particle's position) is evaluated. For this paper, we will summarise the approaches below:

1. $FE_1(Particle_i) = P^m * S^m$
   $S^m$ represents a qubit-state such as "11111". It is also noted that $S^m$ also represents the integer value of the qubit-state's binary string, such as "11111" being 31. $P^m$ represents the probability of the qubit-state $S^m$. Hence, both variables are interlinked. The maximum value given from this fitness evaluation in a 5 qubit state is $31 * 1 = 31$.

2. $FE_2(Particle_i) = \sum_{k=0}^{n-1} P^k * S^k$
    $n$ represents the total number of qubit states, and with a 5-qubit state, of a particle. The remaining part of the function is similar to the previous evaluation function.

[7] provides further details and examples of the stated fitness evaluation approaches. In summary, the first approach focuses on the probability of the qubit state that is the closest to the maximum value of the 5-qubit MaxOne problem. This will disregard the probabilities assigned to other intermediary states. The second approach, however, focuses on a "weighted average" of the experiment iterations which results in the intermediary states having a bigger impact on the results.

In terms of the PSO implementation, the first evaluation approach will prioritise finding the state with the highest probability, effectively emphasising exploitation. In theory, this may lead to a quicker convergence towards solutions with high probabilities. The second approach takes into account a broader spectrum of states, which could enhance exploration and may spend more time exploring different regions of the solution space before converging.

## 5. Experiments and Results Analysis

The main method of experimentation is through running the algorithm implementation various times and noting statistical points of fitness within the graph over time or iterations. The PSO algorithm is written in Python, similar to the GA implemented in [7]. Generally, a large number of iterations are required during experimentation to ensure that the stochastic effect can be better considered (more reliable randomness and evolution), hence this paper runs the algorithm up to $t = 29$ (30 iterations). In every experimentation, we initialised 50 particles and a proceeded with a random mutation to each gates. The results are plotted with Matplotlib [26] and an overview of the worst, average and best fitnesses of the population at each iteration are shown for better clarity. A plot graph of all the particle's fitnesses through the iterations will also be shown for further clarity. We note that the results shown are random instances of experimentation. As there can be many variations of a PSO implementation, this section will be divided into multiple subsections that focus on different aspects of the implementation. Each subsection will present its conclusion based on the results shown. These variations could be explored classically in Section 3.d. The sections are as follows: a comparison of the learning capabilities, a comparison of Inertia Weight variations and a comparison of evolutionary algorithm for quantum circuit synthesis.

a. **A Comparison of Learning Capabilities**
This experiment focuses on comparing the different learning capabilities of the PSO algorithm. It can be done by simply tweaking and slightly modifying the learning parameters within the velocity equation (5), specifically parameters $c_1$ and $c_2$. For this experiment, we take the inertia weight to be a constant 1 throughout the entire span of iterations. The exact parameter values used in this experiment can be seen in **Figure 3**.

| Learning Experiments | $c_1$ | $c_2$ | $w$ |
|---|---|---|---|
| Balanced Learning | 1.5 | 1.5 | 1 |
| Cognitive Learning | 4.0 | 1.5 | 1 |
| Social Learning | 1.5 | 4.0 | 1 |

**Figure 3**

We analyse and compare the different learning capabilities of the algorithm through utilising different constants, affecting the way a particle behaves within an instance. It should also be noted that the inertia weight in this experiment stays constants throughout the experiments. The exact numbers shown were randomly chosen, but we ensured a more biased learning through higher constant values.

Figures 5 and 6 shows a result of this experiment. These two results differ by the use of the two methods of fitness evaluation discussed in Section 4.d. In both fitness evaluations, Social Learning presents to be the best learning method for a particle in this PSO implementation. The particles plot also show that social learning allows a particle to keep it's high fitness evaluation over multiple iterations. This result seems consistent with other instances of experimentation while running the PSO implementation, with some experiments showing the balanced learning setting better fitness values than social learning.

**b. A Comparison of Inertia Weight**

Described previously, the inertia weight of PSO can be used as an experiment to analyse the PSO implementation. This experiment uses two methods of determining the inertia weight utilised in the algorithm, a predefined-constant and a time-varying inertia weight (CIW and TVIW respectively). As the main purpose of this experiment is simply to compare the different methods of introducing the inertia weight. This experiment is inspired by works such as [27], where an inertia weight comparison was also introduced. The defined parameters used in this experiment can be seen in **Figure 4**.

| Inertia Weight Experiments | Initial $w$ ($w_1$) | Final $w$ ($w_2$) |
|---|---|---|
| Predefined-Constant IW | 1 | 1 |
| Time-Varying IW | 1 | 0.3 |

**Figure 4**

In an attempt to improve PSO, [27] introduced a Time-Varying Inertia Weight where the inertia weight linearly decreases over iterations. Generally, the idea is that an initial stage of large inertia weight would enhance a search of new area (global exploration) and over time, lower inertia weight for a local exploration. This experiment uses the mathematical expression showcased in [27], whilst utilising the values described in Figure 3. For clarity, the mathematical expression is described as below:

$$\text{Inertia } w_t = (w_1 - w_2)(\frac{t_{max} - t}{t}) + w_2$$

$t_{max}$ represents the total amount of iterations and $w_t$ represents the inertia weight utilised at iteration $t$. The exact value decreases over time, and in this experiment the final inertia weight ($w_2$) is set to be 0.3. For ease of comparison, the algorithm will utilise a Social Learning bias such that the focus of analysis goes towards the inertia weight comparison. Results for this experimentation can be found in Figure 7 and 8, where time-varying inertia weight noting a higher fitness value. However, throughout multiple instances of experimentation, predefined constant-inertia weight results in a higher fitness value, as shown in Figure 9.

c.  **General Comparison of GA and PSO for Quantum Circuit Synthesis**

This section will focus on an analytical comparison of the PSO algorithm proposed in this paper with a Genetic Algorithm proposed in [7] for quantum circuit generation. It should be noted that both generated quantum circuits from both algorithms are to solve the MaxOne problem. Hence, there are little to no difference in the fitness evaluation technique described in both implementations. There are a few literal differences that can be pointed out, one of which is the encoding methods of the algorithm and its impact on diversity.

The genetic algorithm encodes quantum gates as genes by grouping each gate type separately and storing them in a list. This method of encoding the gates is similar to how the particle swarm optimisation approach represents quantum gates - as particles in a list. However, the grouping of each type of component is more impactful and practical in the genetic algorithm due to the genetic mutation that can happen to all aspect of a quantum gate. This allows for an easy exchange between different chromosomes without affecting the values of the gates, impacting the diversity in the genetic makeup of the population. The PSO implementation utilises the grouping of gates in its random mutation phase to determine the next position of a particle. This will also impact the diversity of the population rather significantly as a more stochastic approach is taken, which then widens the solution search space.

Another comparison that could be done is to evaluate the convergence rate of both implementations. The convergence rate can be determined by analysing its performance over multiple iterations or epochs. The results of experimentation are plotted over time with regards to its objective fitness. The results, such as Figure 5, show that the PSO implementation converges to the best fitness value (over a specific run) faster than the GA implementation. This observance cannot be substantially proven due to a difference in seed and randomness, but throughout various experimentation, it could be taken that the PSO implementation converges to the best fitness (given a population) faster than the GA.

## 6. Conclusion

We have explored a particle swarm optimisation (PSO) approach for synthesizing quantum circuits to solve the maxone problem. The key elements of the explored of this PSO implementation are the method of encoding quantum circuit instructions as particles, defining position and velocity representations, and evaluating circuit fitness. Experimentation was conducted to compare different learning capabilities by adjusting the cognitive and social learning parameters. Results showed that social learning enabled particles to maintain high fitness over iterations, with additional testing to analyse various parameters effects.

Experiments showed that a higher social learning weight, which relies more on the global best solution, enabled particles to maintain higher fitness over iterations compared to balanced or cognitive learning modes. This indicates that social learning is most effective for this problem. Additionally, using a time-varying inertia weight that decreases over iterations appeared to improve optimization and balance of global and local search compared to a constant weight. Finally, benchmarking the PSO against a previous genetic algorithm method revealed that the PSO approach converged to optimal solutions faster, likely due to increased population diversity from the stochastic mutation phase. This faster convergence could potentially indicate it became stuck at a local maxima. No optimization techniques were implemented to prevent the PSO from getting trapped at local maxima rather than continuing to explore the global search space. The faster convergence may simply reflect that the PSO exploited solutions in a limited region, rather than thoroughly exploring diverse areas of the landscape. Further testing with optimization methods enabled would be needed to determine if the PSO can escape local maxima and reliably find the global optimum.

In conclusion, this work presented a novel PSO algorithm for automated quantum circuit synthesis. The performance was analyzed through variations in learning parameters, inertia weight strategies, and benchmarking to an alternative technique. Outcomes indicate the PSO approach can effectively navigate the search space to generate quantum circuits solving specified problems. Future work could enhance the encoding or fitness evaluation and compare convergence speeds over larger sample sizes.

# 7. Appendix
## Figures

**Overview**

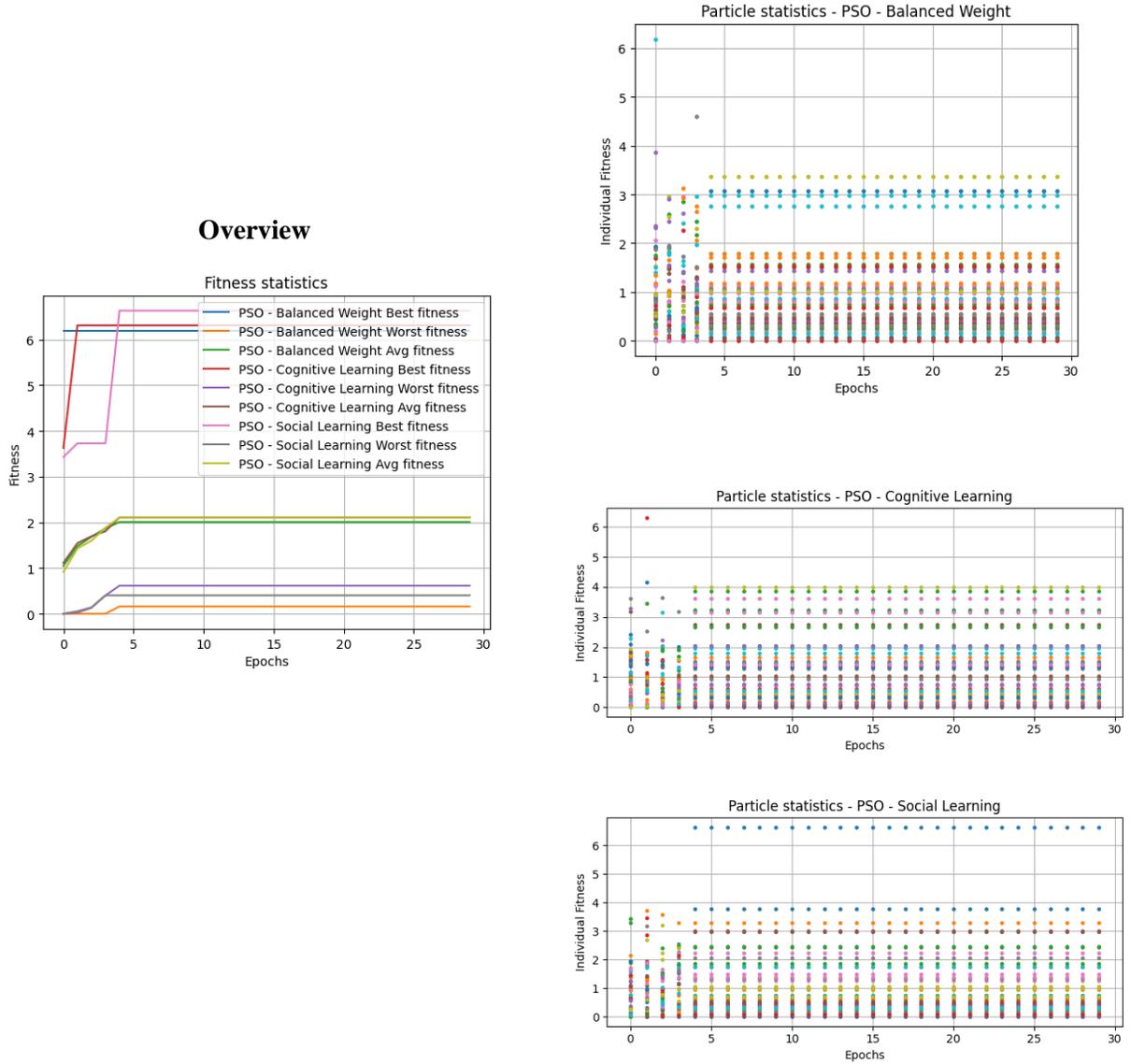

**Figure 5 - Learning Experiments & $FE_1$**

## Overview

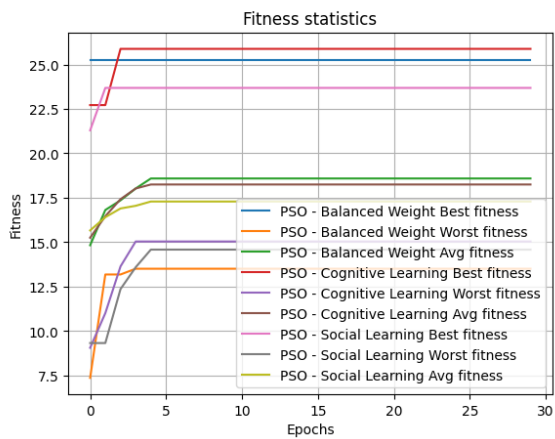

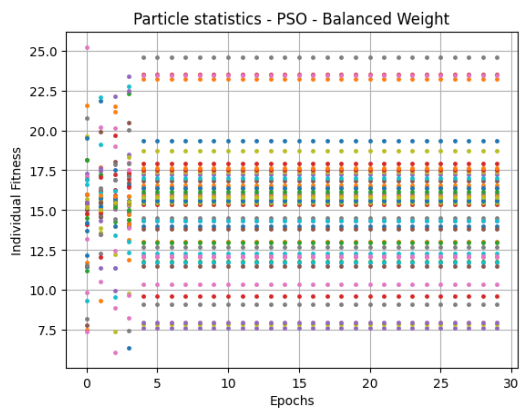

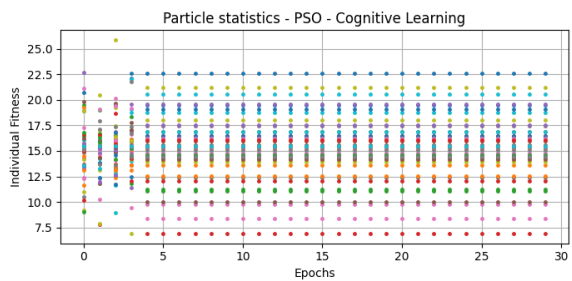

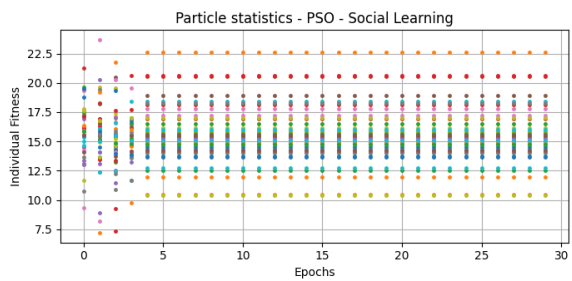

**Figure 6 - Learning Experiments & $FE_2$**

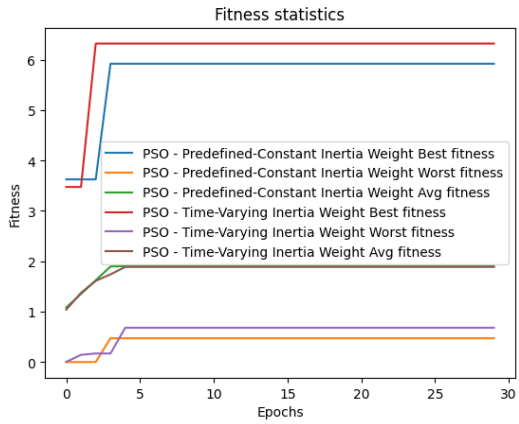
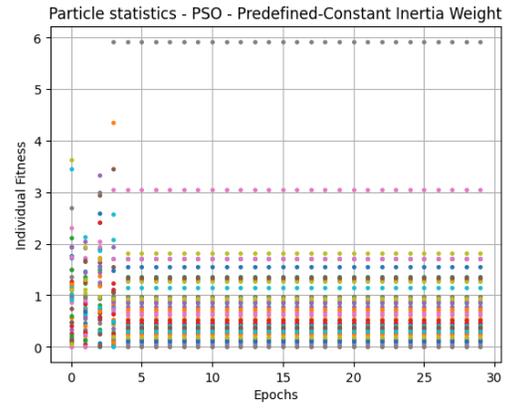
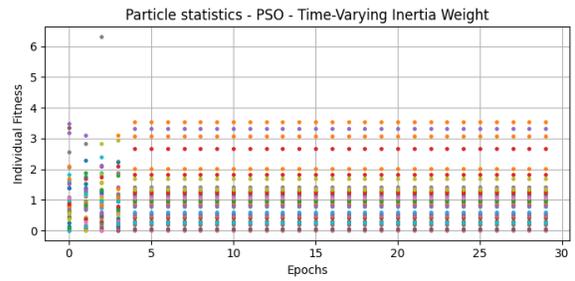

**Figure 7 - Inertia Weight Experiments & $FE_1$**

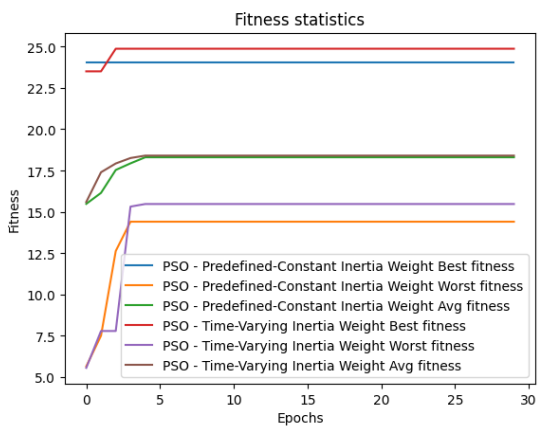
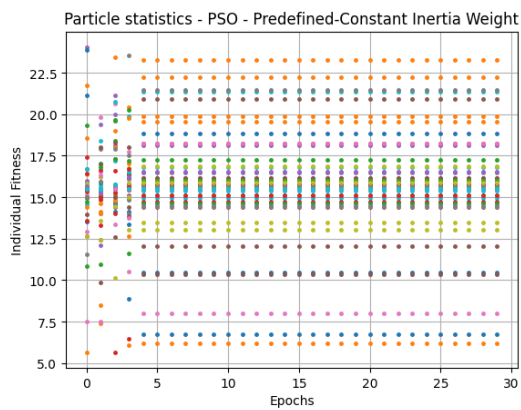
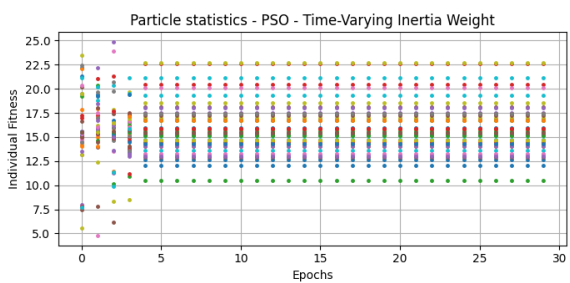

**Figure 8 - Inertia Weight Experiments & $FE_2$**

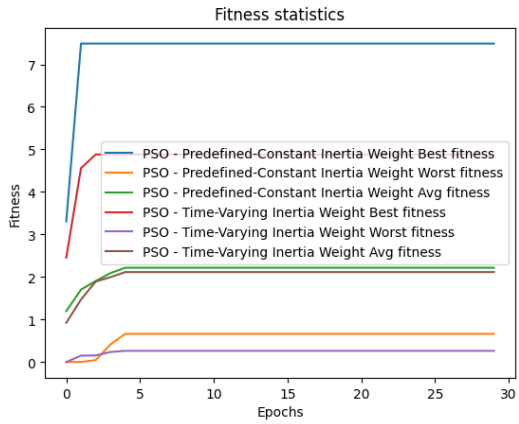
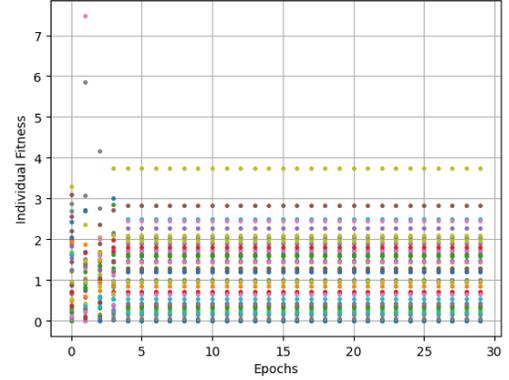
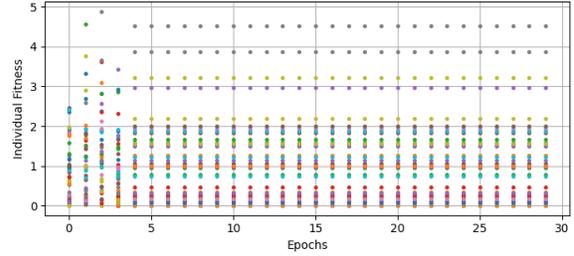

**Figure 9 - Inertia Weight Experiments & $FE_1$**